\documentclass{article} % For LaTeX2e
\usepackage{times}
\usepackage{hyperref}
\usepackage{url}
\usepackage{subcaption}

\usepackage{amsfonts}  % for mathbb
\usepackage{graphicx}
\usepackage{caption}
\usepackage{amsmath}

\usepackage{wrapfig}
\usepackage{amssymb,amsmath,amsthm,mathtools}

\usepackage{appendix}
\usepackage{etoolbox}
\usepackage{lipsum}
\usepackage{natbib}

\usepackage{xr}

\usepackage[margin=1in]{geometry}
\makeatletter
\renewenvironment{abstract}{%
    \if@twocolumn
      \section*{\abstractname}%
    \else %% <- here I've removed \small
      \begin{center}%
        {\bfseries \Large\abstractname\vspace{\z@}}%  %% <- here I've added \Large
      \end{center}%
      \quotation
    \fi}
    {\if@twocolumn\else\endquotation\fi}
\makeatother

\usepackage{cleveref}

\newcommand{\sectionname}[0]{Section~}
\renewcommand{\figurename}[0]{Figure~}

\DeclareMathOperator*{\argmin}{arg\,min}

\title{Negative eigenvalues of the Hessian\\ in deep neural networks}

\author{Guillaume Alain \thanks{This work was done during an internship with the Google Brain team in Montreal.} \\
Mila, University of Montreal
\and
Nicolas Le Roux \\
Google Brain
\and 
Pierre-Antoine Manzagol \\
Google Brain
}

\newcommand{\paperroot}{.}

\begin{document}

\maketitle

\begin{abstract}
The loss function of deep networks is known to be non-convex but the precise nature of this nonconvexity is still an active area of research. In this work, we study the loss landscape of deep networks through the eigendecompositions of their Hessian matrix. In particular, we examine how important the negative eigenvalues are and the benefits one can observe in handling them appropriately.
\end{abstract}

\section{Introduction and related work}

Although deep learning has had many successes, the optimization of deep models remains difficult and slow. One of the main reasons is that the loss landscape of such networks is non-convex. While there is a good understanding of the optimization of convex functions~\citep{bottou2018optimization}, or even specific non-convex functions such as PCA~\citep{de2014global}, the theory about general non-convex functions is still poorly understood despite a lot of recent theoretical progress~\citep{tripuraneni2017stochastic,allen2017natasha}. Most of these last advances focus on dealing with saddle points, either through random perturbations of the gradient or cubic regularization~\citep{nesterov2006cubic}.

This focus on saddle point stems in part from past analyses detailing their omnipresence in the loss landscape of deep networks~\citep{dauphin2014identifying, choromanska2015loss}, as well as the fact that many local minima are of such high quality that we do not need to worry about not having the global minimum. Although explicitly handling saddle points is appealing, it has its own issues. In particular, most of these methods need to solve an inner-loop when close to a saddle point, either to find the direction of most negative curvature~\citep{allen2017natasha} or to solve the cubic regularized problem~\citep{tripuraneni2017stochastic}. This increases the practical complexity of these methods, potentially limiting their use.

The Hessian of the loss has been the topic of many studies. In addition to the work of~\citet{dauphin2014identifying},~\citet{papyan2018eigenvalues} studied the spectrum of the Hessian and other matrices at a stationary point while~\citet{sagun2016eigenvalues} analyzed the evolution of the spectrum of the Hessian during the optimization.~\citet{gurari2018tiny} went further and also analyzed the eigenvectors. In particular, they showed that the Hessian  was almost low-rank and that the subspace spanned by the top eigenvectors remained relatively stable during optimization.

Having as ultimate goal to design efficient optimization methods for deep networks, we focus on studying properties of the loss that would affect such methods and explore several questions.

First, we study how quickly the Hessian changes during optimization. Rather than tracking the top subspace as done by~\citet{gurari2018tiny}, we compute the top and bottom eigenvectors of the Hessian at a given point and track the curvature in these directions. The goal is to assess whether second-order methods which slowly update their approximation of the Hessian can correctly capture the current geometry of the loss.

Second, we explore the accuracy of the second order approximation of the loss. We observe that, while this approximation closely matches the true loss in directions of positive curvature, this is far from the case in the directions of negative curvature. This raises the question of the scale at which we should build this approximation for efficient optimization.

Third, we focus on the directions of negative curvature. We study how much of the potential decrease in training loss is contained in these directions of negative curvature, hoping to understand how important it is to design optimizers which can make use of them. We also study the relationship between the curvature $\rho$ in a direction and the stepsize $\alpha^*$ maximizing the gain in that direction. We find that while we roughly have $\alpha^* = 1 / |\rho|$ in directions of positive curvature, this relationship does not hold in the directions of negative curvature.

\section{A curvature analysis}
\subsection{Experimental setup}
\label{neh-sec:methodology}

With the flurry of architectures, optimization methods and datasets currently available, every experimental study is bound to be incomplete. Ours is no exception and we focus on one architecture, one optimizer and one dataset. While we make no claim that our results remain valid across all possible combinations, we believe that them being true for the combination tested already offers some insights.

Another limitation of our results is our focus on the training loss. We know the relationship between the training loss and the generalization loss to depend on all three aspects mentioned above~\citep{zhang2016understanding}. While a discussion on the joint properties of the training and generalization loss would definitely be of interest, it is outside the scope of this study.

\paragraph{Architecture:} We used a LeNet architecture~\citep{lecun1989backpropagation} with ReLU as activation function. It has two convolutional layers, two fully connected layers, and a softmax on the last layer, for a total number of approximately $d=3.3\times 10^6$ parameter coefficients. While this is a network of reasonable size, far larger networks exist, some of them built specifically to make optimization easier~\citep{he2016deep}.

\paragraph{Optimizer:} To compute an optimization trajectory, we used RMSProp~\citep{hinton2012neural} with a batch size of 32, and an initial learning rate of 0.00036 with exponential decay at every training step for a combined decay of 0.75 at every epoch. The RMSProp decay rate is 0.95 and its momentum is 0.22.

\paragraph{Dataset:} We performed experiments with MNIST \citep{lecun1998mnist}. While there have been attempts at estimating Hessians in very high dimensions, for instance by~\citet{adams2018estimating}, they tend to suffer from either even higher computational costs or large variance. %Hence, it is an open question whether our conclusions will extend to larger architectures or datasets, although results by~\citet{papyan2018eigenvalues} suggest they do.

Since all the eigenvalues are real-valued because of the symmetry of the Hessian, they can be ordered as $\lambda_1 \geq \lambda_2 \geq \ldots \geq \lambda_d$. %See Appendix \sectionname\ref{neh-appsec:jvp} for details on how we can compute the $k$ largest or smallest eigenpairs $(\lambda_i, v_i)$.

\subsection{Tracking Hessian eigenvectors through time}
\begin{figure*}[ht!]
% \centering
%\includegraphics[width=.48\textwidth]{\paperroot/figures/eig_M/eigenvalues_over_time_split00of20_logscale.png}
%\includegraphics[width=.48\textwidth]{\paperroot/figures/eig_C/eigenvalues_over_time_split00of20_logscale.png}
\includegraphics[width=.48\textwidth]{\paperroot/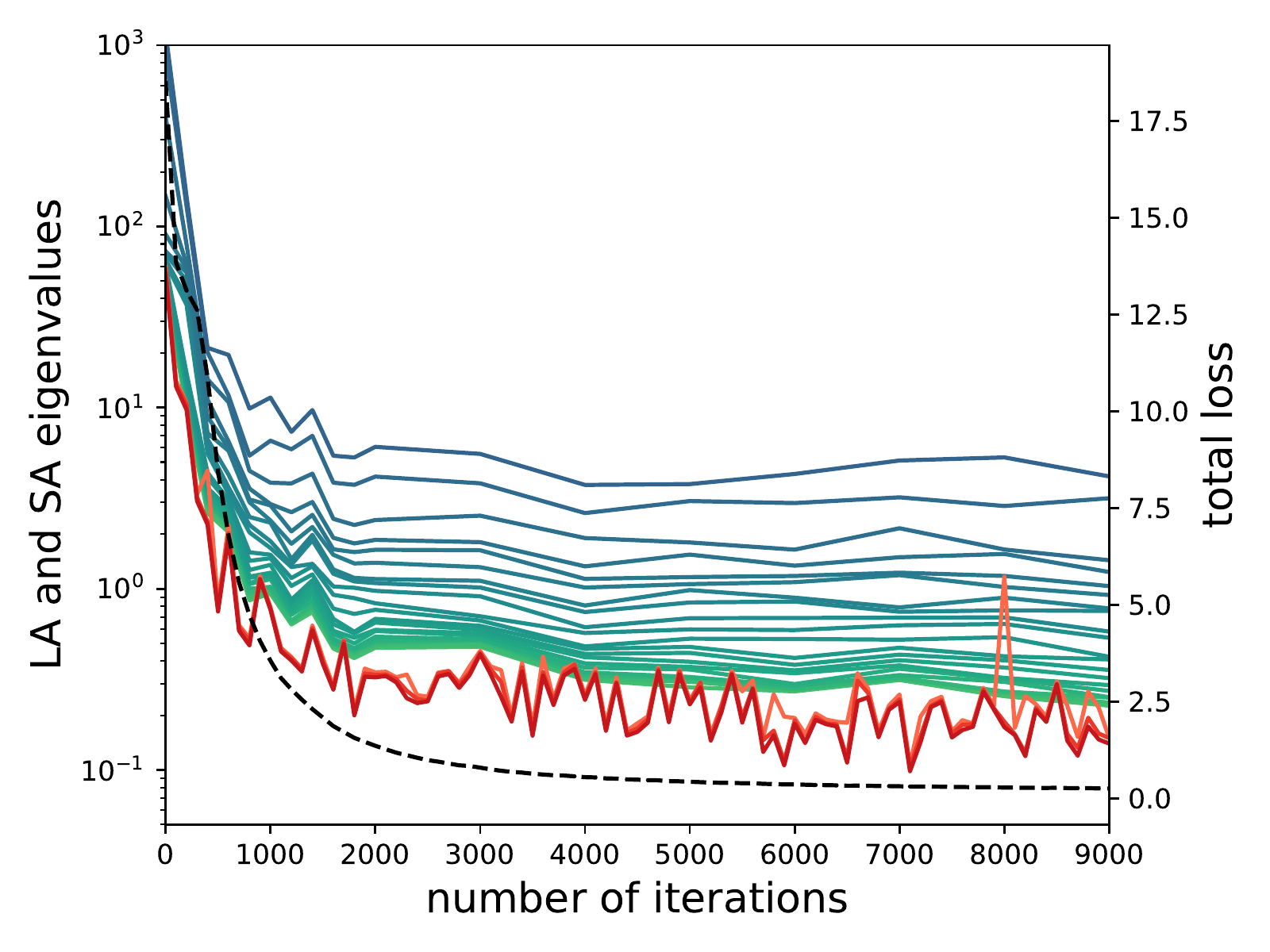}
\hspace{0.02\textwidth}
\includegraphics[width=.48\textwidth]{\paperroot/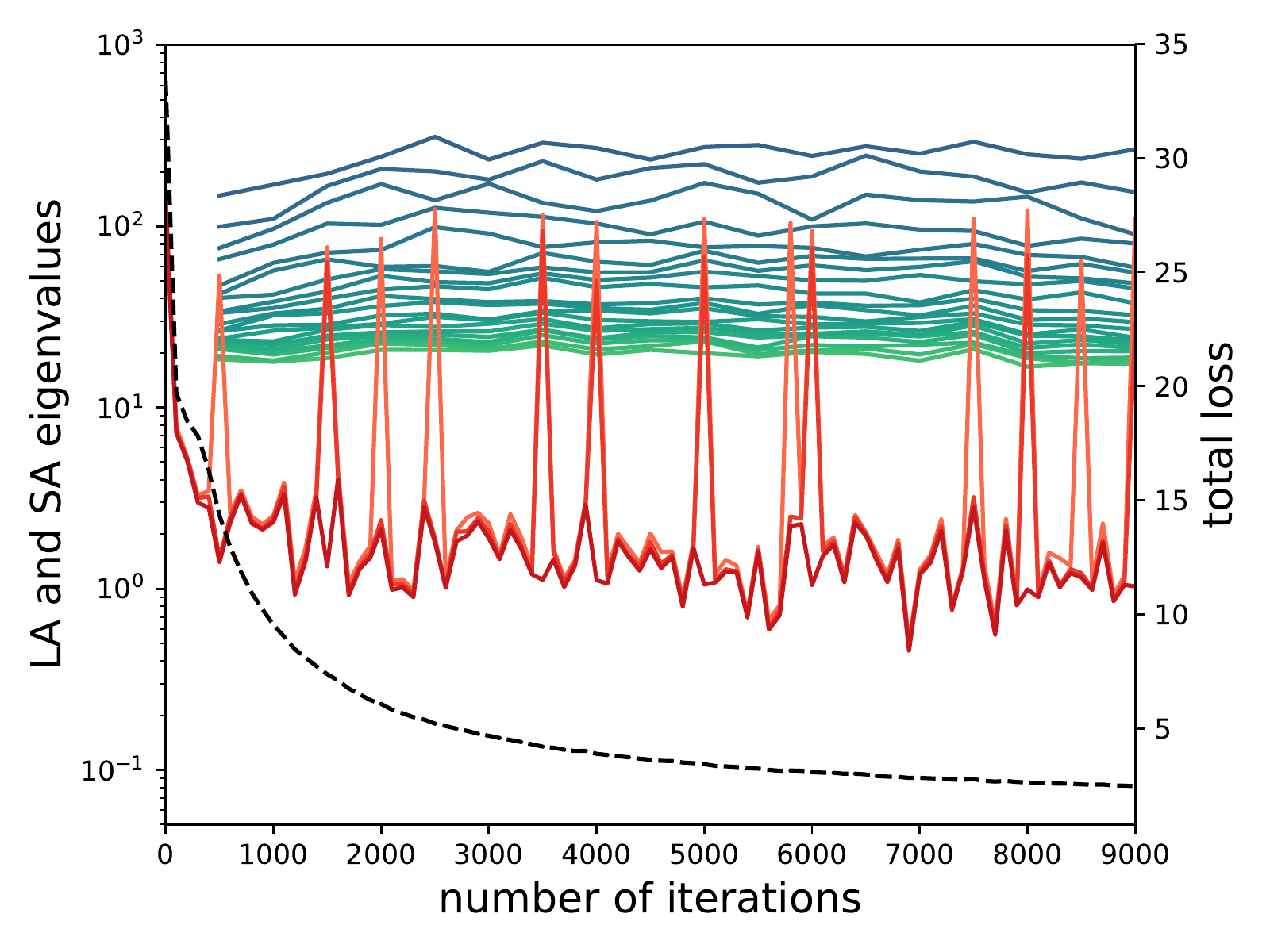}
\caption{Evolution of the logarithm of the absolute eigenvalues of Hessian during training on the MNIST (left) and CIFAR-10 (right) datasets. Largest positive eigenvalues are in blue/green, largest negative eigenvalue is in red. We see that the largest positive eigenvalues stabilize after a while on both datasets. The behaviour of the largest negative eigenvalue is dataset dependent. The dotted black curve is the total training loss.
\label{neh-fig:eig_imperfect_eigvals_100}}
\end{figure*}

It is well-known that, in the noiseless strongly convex case, the convergence rate of the excess error when using a first-order algorithm is linear with a speed that depends on the condition number of the Hessian, i.e. the ratio of largest to smallest eigenvalue. One can view the condition number as a ``spatial spread'' in that it describes the spread of eigenvalues in different directions. Reducing that spread, and thus increasing convergence, can be done by preconditioning the parameters, i.e. by updating the parameters using
\begin{align*}
    \theta_{t+1} = \theta_t - \alpha M g_t \; .
\end{align*}
In the noiseless quadratic case, the optimal preconditioner is $M = H^{-1}$, giving rise to Newton method. The use of $H^{-1}$ reduces the spread from the condition number to 1, effectively achieving convergence in one step.

If the function is not quadratic, one can still use the Hessian computed at one point to precondition the gradients at another point. In that case, the convergence rate will not depend on the condition number of the Hessian at the current point but rather at the discrepancy between the Hessian at the current point and the Hessian used for preconditioning. In other words, it is the variation of the Hessian through time which impacts the convergence, which we denote ``temporal spread''\footnote{This temporal spread depends both on the third derivative of the function and the distance travelled in parameter space.}.

Figure~\ref{neh-fig:eig_imperfect_eigvals_100} shows the evolution of the largest positive and negative eigenvalues of the Hessian during training on both the MNIST and CIFAR-10 datasets. These results resemble those of~\citet{dauphin2014identifying} and~\citet{gurari2018tiny}, among others. From that figure, it might seem the Hessian stabilizes after a few iterations and second-order methods should be efficient, even when the approximation to the Hessian is built over many timesteps to reduce the noise.

\begin{figure*}[ht!]
\begin{subfigure}[t]{0.48\textwidth}
    \centering
    \includegraphics[width=\textwidth]{\paperroot/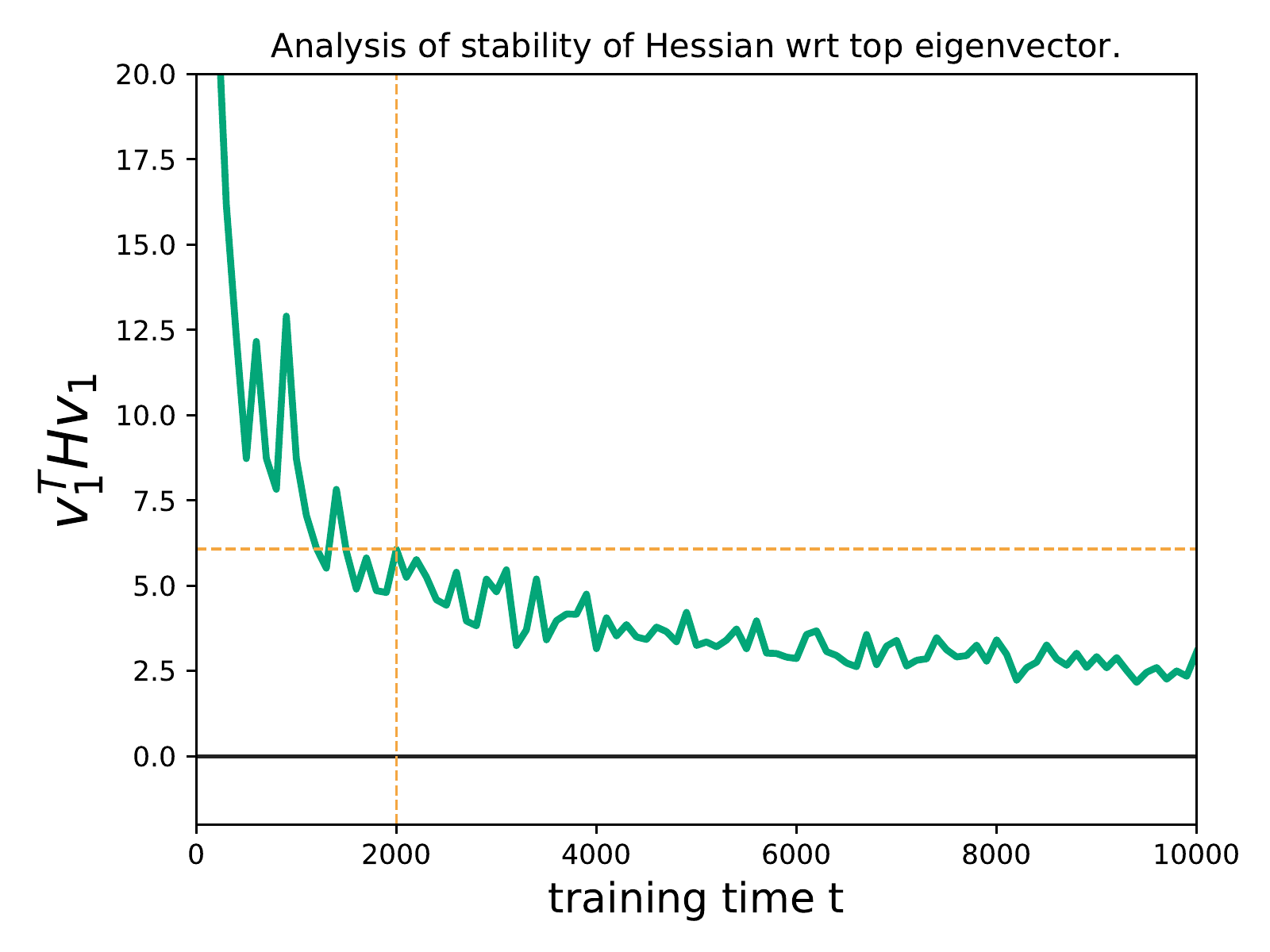}
    \caption{Top eigenvector at $t_0=2000$}
    \label{neh-fig:top_eig_vs_hessian_over_time_2000}
    %\vspace{2em}
\end{subfigure}
\begin{subfigure}[t]{0.48\textwidth}
    \centering
    \includegraphics[width=\textwidth]{\paperroot/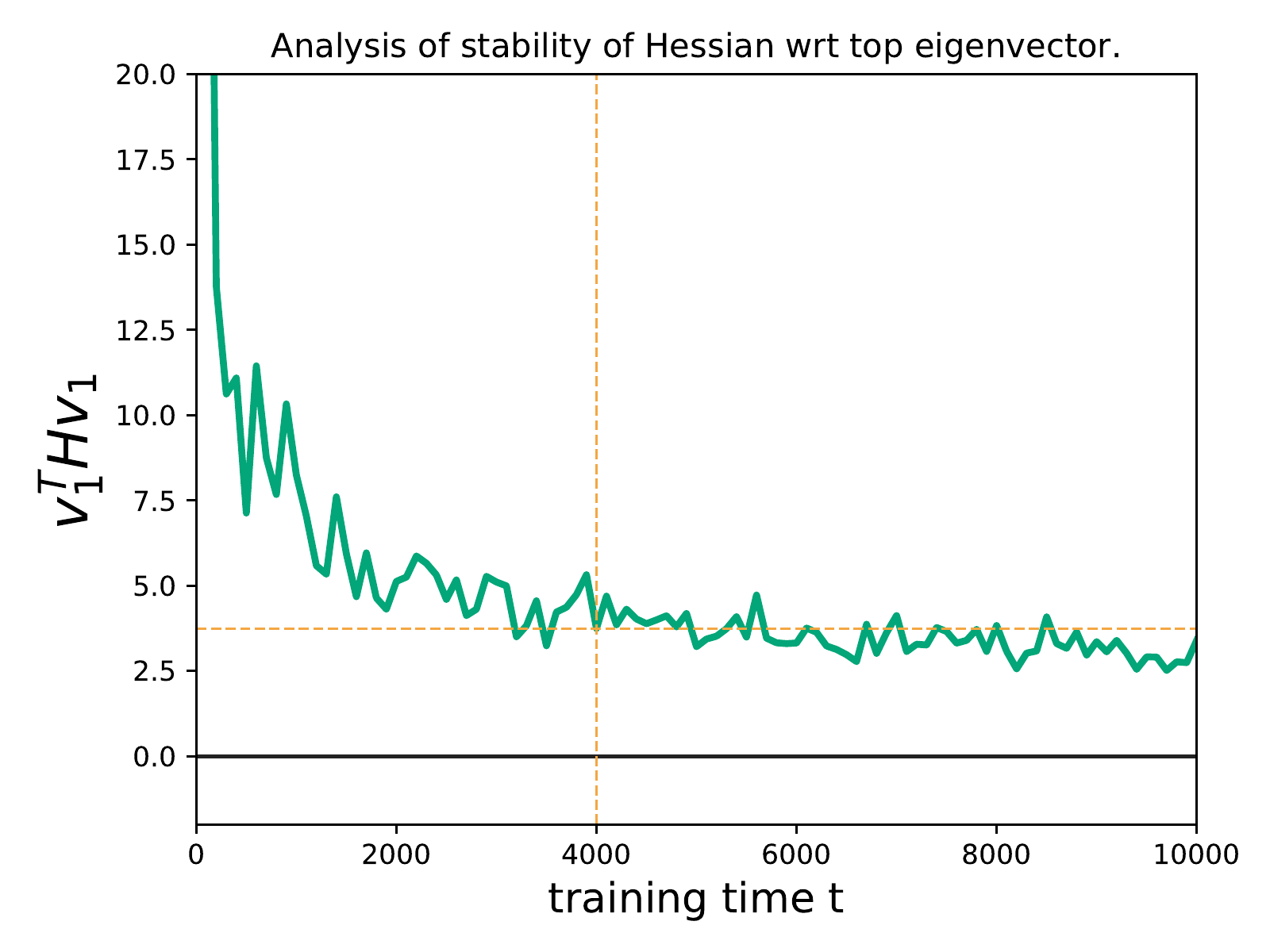}
    \caption{Top eigenvector at $t_0=4000$}
    \label{neh-fig:top_eig_vs_hessian_over_time_4000}
    %\vspace{2em}
\end{subfigure}
\begin{subfigure}[t]{0.48\textwidth}
    \centering
    \includegraphics[width=\textwidth]{\paperroot/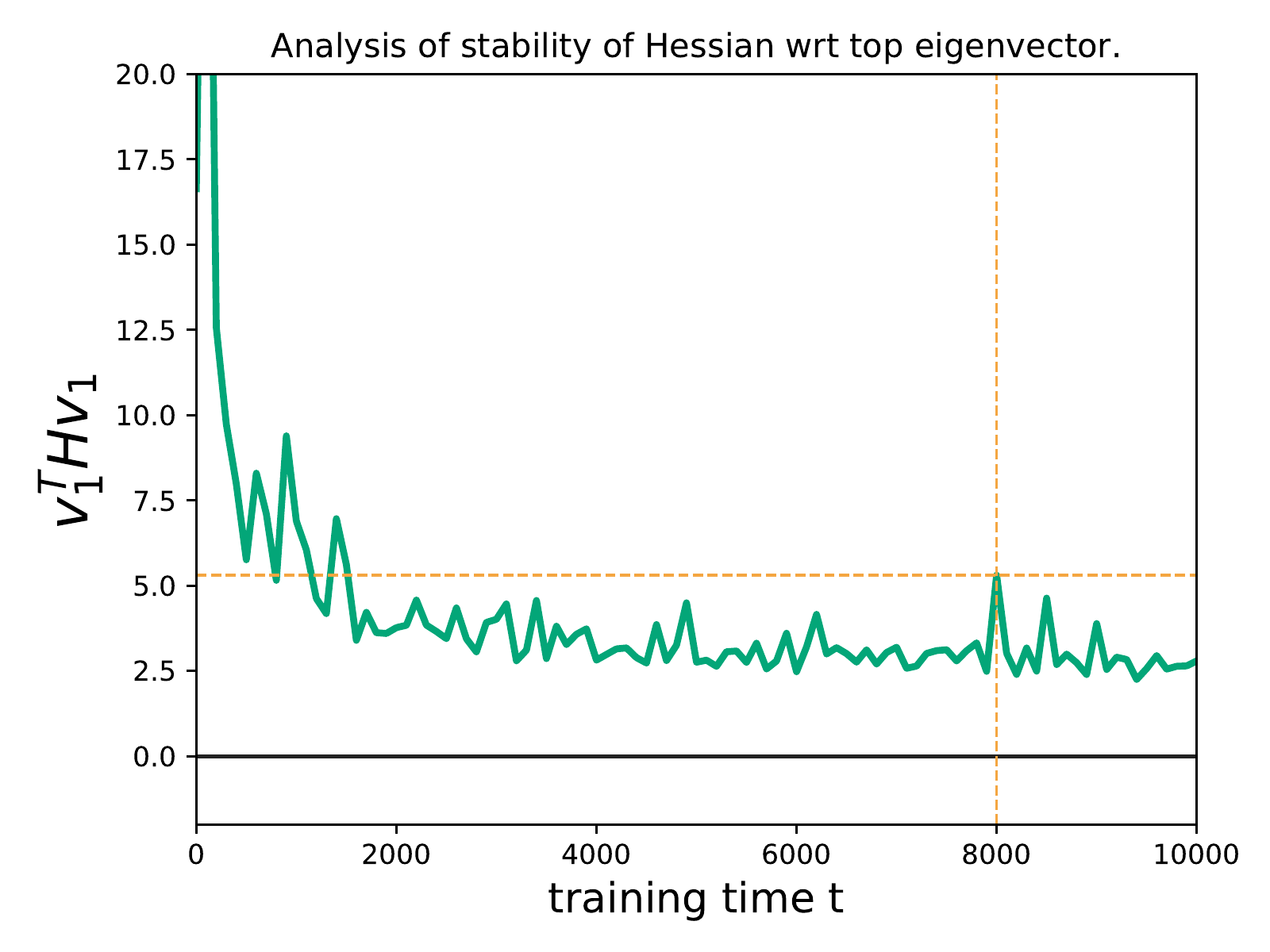}
    \caption{Top eigenvector at $t_0=8000$}
    \label{neh-fig:top_eig_vs_hessian_over_time_8000}
    %\vspace{2em}
\end{subfigure}
\begin{subfigure}[t]{0.48\textwidth}
    \centering
    \includegraphics[width=\textwidth]{\paperroot/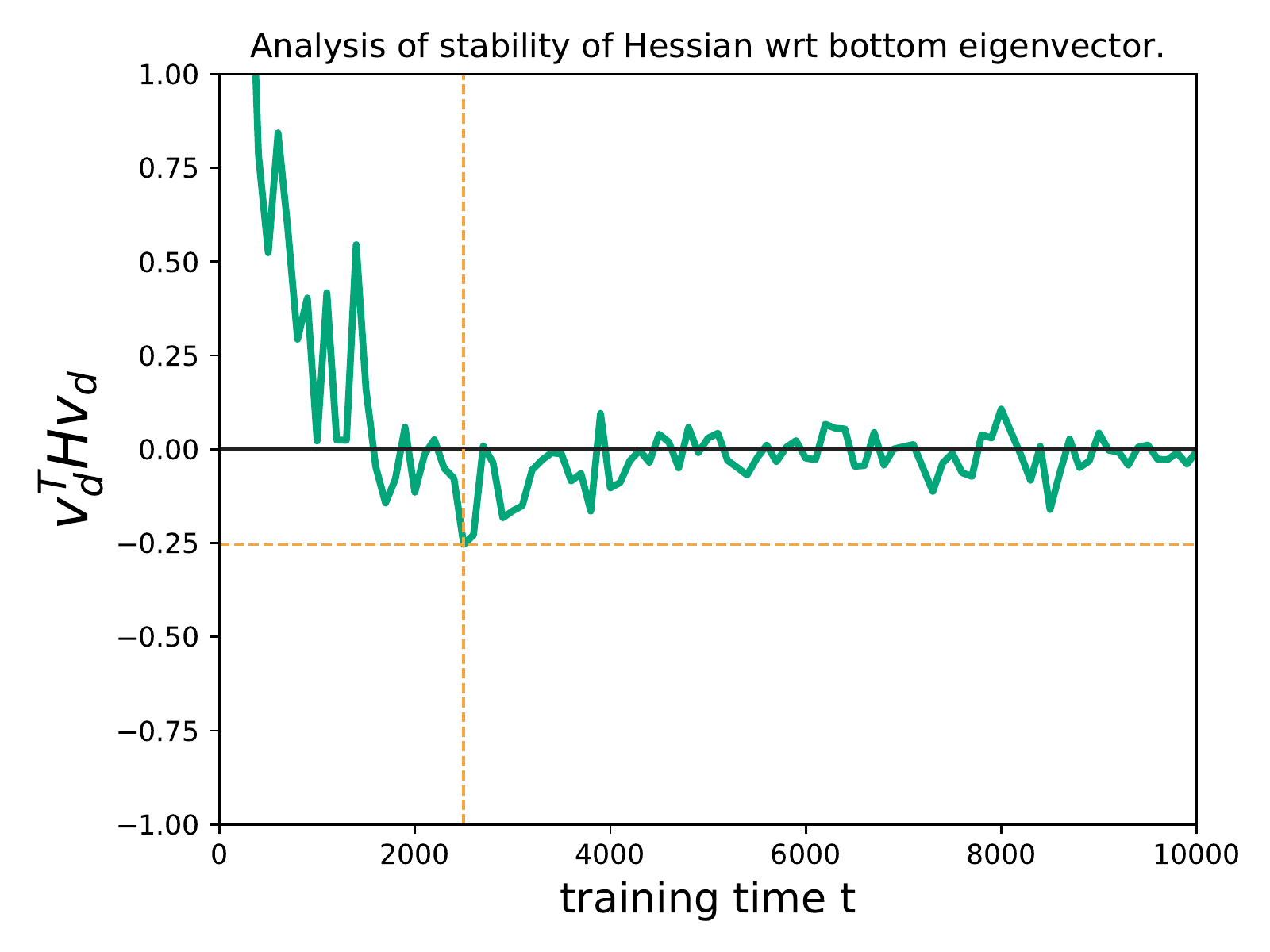}
    \caption{Bottom eigenvector at $t_0=2500$}
    \label{neh-fig:bottom_eig_vs_hessian_over_time}
    %\vspace{2em}
\end{subfigure}
\caption[Eig vs Hessian]{Comparing the directions of curvature evolving over time for top eigenpair at $t_0=2000$ (top left), $t_0=4000$ (top right), $t_0=8000$ (bottom left) and bottom eigenpair at $t_0=2500$ (bottom right). For the given eigenvector $v$ computed at those specific moments in time, we compare the values of $v^\top H(t)v$ over the range of all Hessians in the training trajectory. Orange lines are traced to help identify the specific time $t_0$ a which the eigenpairs were computed.}
\label{neh-fig:eig_vs_hessian_over_time}
\end{figure*}

It it possible, however, for the Hessian to change despite the spectrum being stable. In that case, second-order methods which slowly update their approximation to the Hessian will be less efficient. To assess the significance of that temporal spread, we compute the top eigenpair $(\lambda_1, v_1)$ at different stages $t_0$ of the optimization, then we plot the curvature $v_1(t_0)^\top H(t) v_1(t_0)$ in that direction for all values of $t$. That way, we can observe whether, even though the spectrum itself is stable, the associated eigenvectors change, making efficient preconditioning difficult.\footnote{Those two particular plots were hand-picked by visual inspection, but they are good representatives of the general trend.}

\Cref{neh-fig:top_eig_vs_hessian_over_time_2000,neh-fig:top_eig_vs_hessian_over_time_4000,neh-fig:top_eig_vs_hessian_over_time_8000} shows the evolving values of $v_1(t_0)^\top H(t) v_1(t_0)$ for $t_0=2000$, $t_0=4000$ and $t_0=8000$ for all values of $t$ in $[0, 10000]$. We see that the curvature in the direction $v_1(t_0)$ follows the same trend as the top eigenvalue of the Hessian, indicating that the top eigenspace of the Hessian is indeed stable, in line with the analysis of~\citet{gurari2018tiny}.

Figure \ref{neh-fig:bottom_eig_vs_hessian_over_time} presents the same analysis using the bottom eigenvector $v_d$ at $t_0=2500$. While the curvature in this direction is negative at $t_0$, it is positive for many other values of $t$. In particular, it varies far more than the smallest eigenvalue of the Hessian shown in Figure~\ref{neh-fig:top_eig_vs_hessian_over_time_2000}. This suggest that the subspace defined by the negative eigenvalues is less stable than the one defined by the positive ones.

In conclusion, while it appears feasible to estimate the subspace spanned by the largest eigenvectors of the Hessian over many timesteps, the same cannot be said to the subspace spanned by the smallest eigenvectors.

Having observed that the Hessian changes throughout the optimization trajectory, we might wonder how accurate the quadratic approximation is. We explore this idea in the next section.

\subsection{Locality of negative curvature}
\label{neh-sec:local_negative_curvature}

\begin{figure*}[ht!]
% \centering
\begin{subfigure}[t]{0.48\textwidth}
    \centering
    \includegraphics[width=\textwidth]{\paperroot/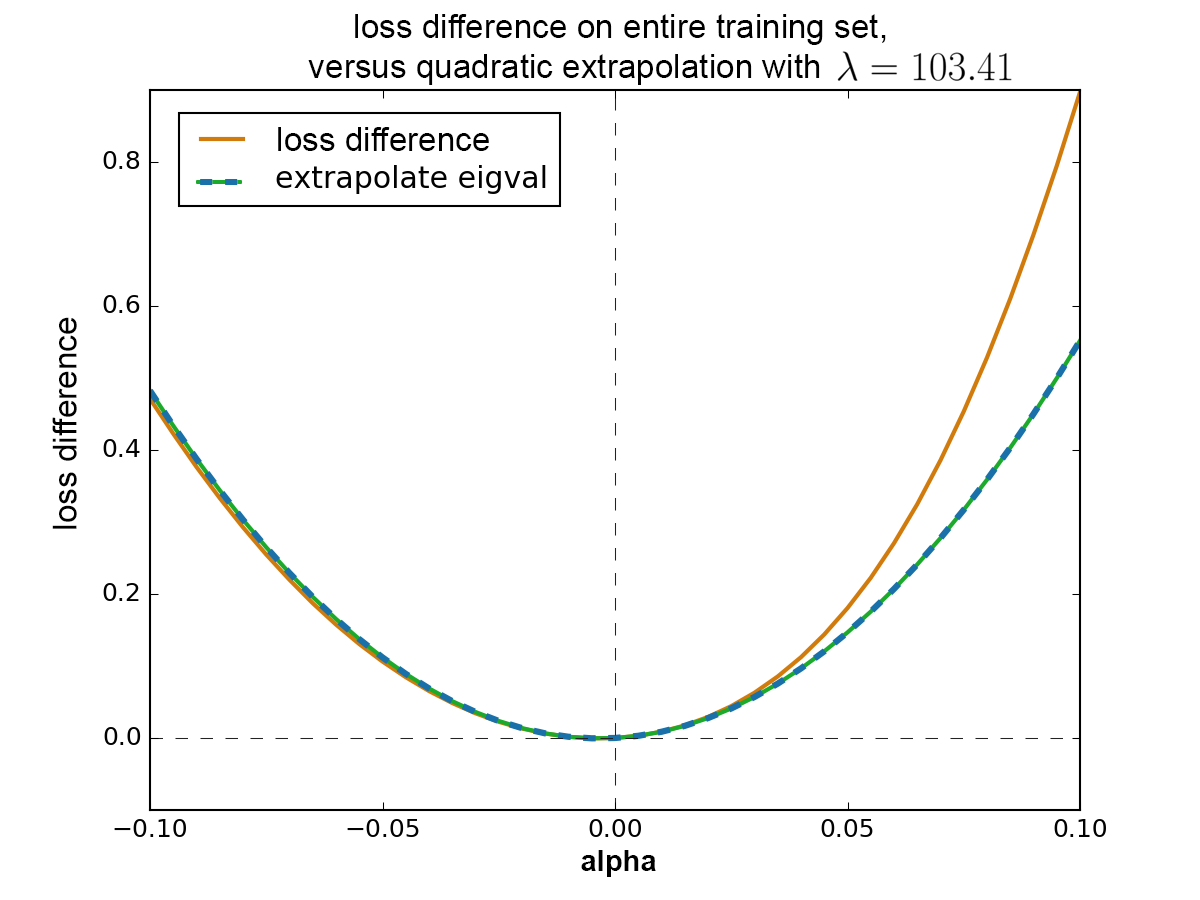}
    \caption[Actual loss vs quadratic approximation, zoomed in]{Top eigenvalue for $\alpha \in [-0.1, 0.1]$}
    \label{neh-fig:measured_diff_vs_expolated_step_top_0p10}
\end{subfigure}
\begin{subfigure}[t]{0.48\textwidth}
    \centering
    \includegraphics[width=\textwidth]{\paperroot/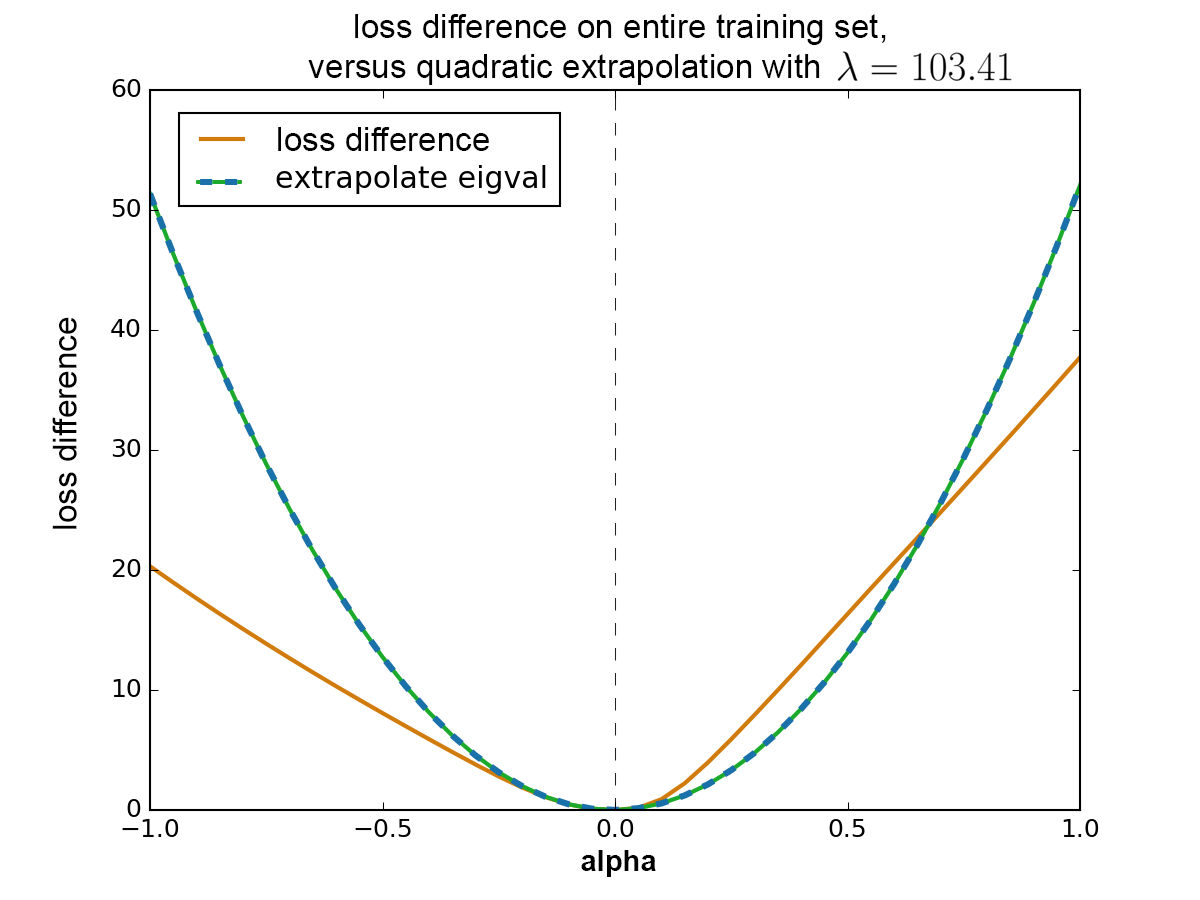}
    \caption{Top eigenvalue for $\alpha \in [-1, 1]$}
    \label{neh-fig:measured_diff_vs_expolated_step_top_1p00}
    \vspace{2em}
\end{subfigure}
\begin{subfigure}[t]{0.48\textwidth}
    \centering
    \includegraphics[width=\textwidth]{\paperroot/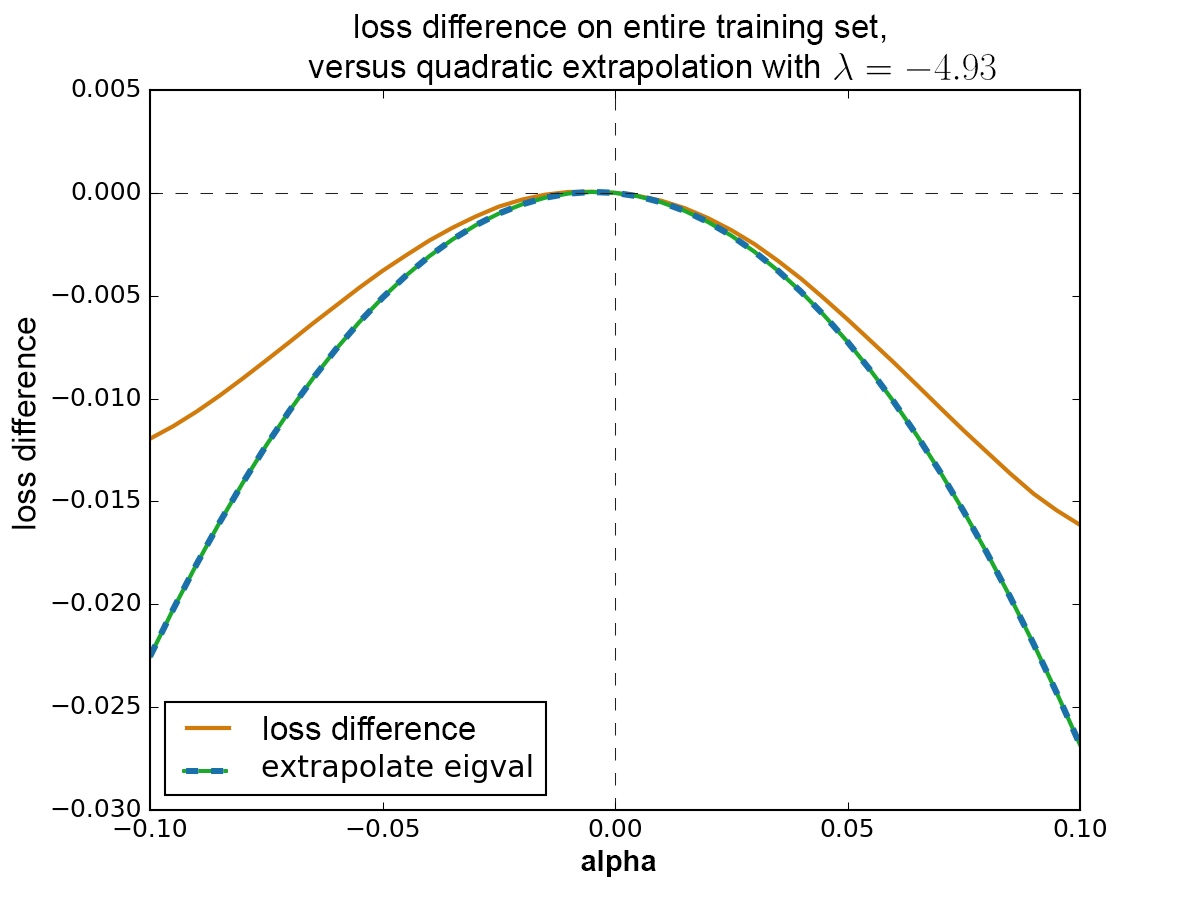}
    \caption[Actual loss vs quadratic approximation, zoomed in]{Bottom eigenvalue for $\alpha \in [-0.1, 0.1]$}
    \label{neh-fig:measured_diff_vs_extrapolated_step_0p10}
\end{subfigure}
\begin{subfigure}[t]{0.48\textwidth}
    \centering
    \includegraphics[width=\textwidth]{\paperroot/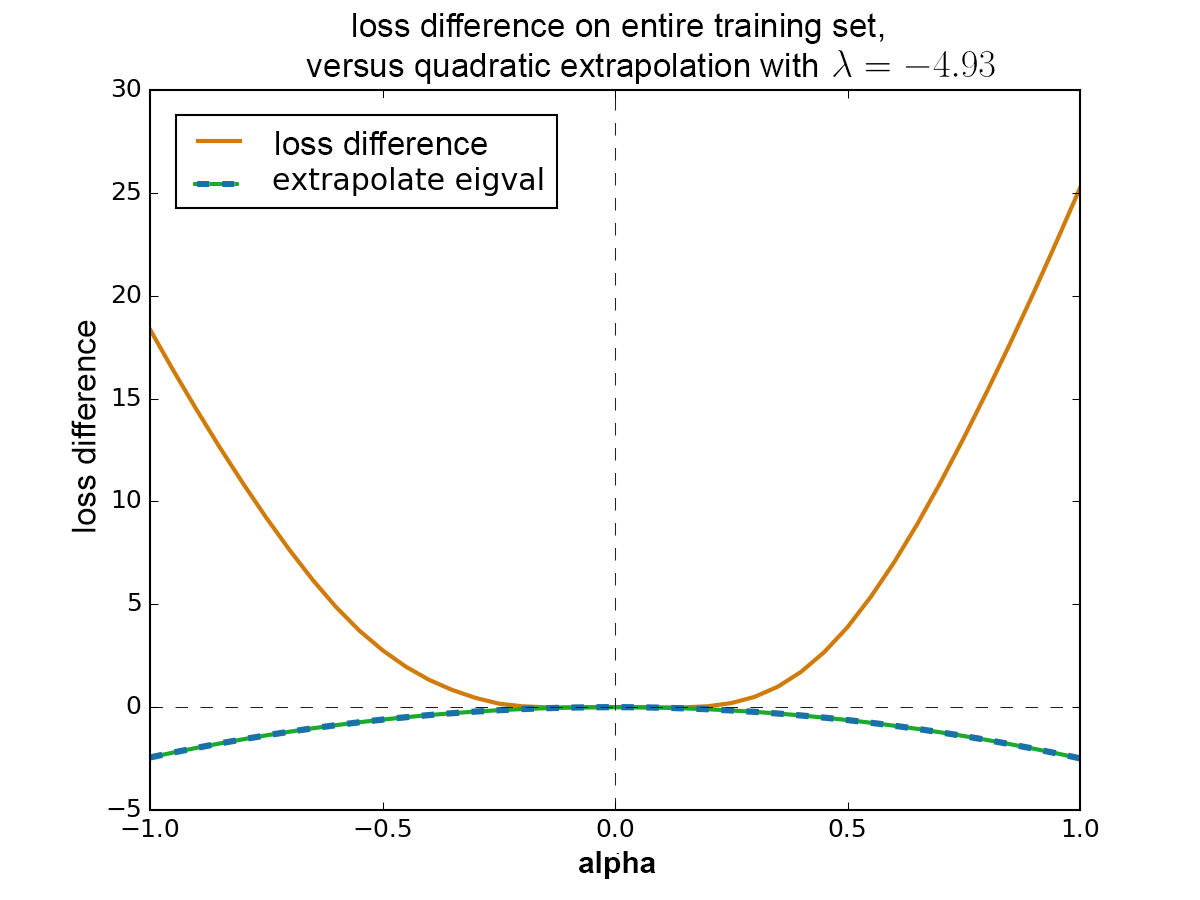}
    \caption{Bottom eigenvalue for $\alpha \in [-1, 1]$}
    \label{neh-fig:measured_diff_vs_extrapolated_step_1p00}
    \vspace{2em}
\end{subfigure}
\caption{Comparison between the true loss and the quadratic approximation for a deep network at small (left) and large (right) scale. We see that, in the direction of largest positive curvature, the quadratic approximation is accurate even for large values of the stepsize $\alpha$. In the direction of largest negative curvature, however, the approximation is wildly inaccurate for larger values of $\alpha$.}
\label{neh-fig:measured_diff_vs_expolated_step}
\end{figure*}

\begin{figure*}[ht!]
\includegraphics[width=0.48\textwidth]{\paperroot/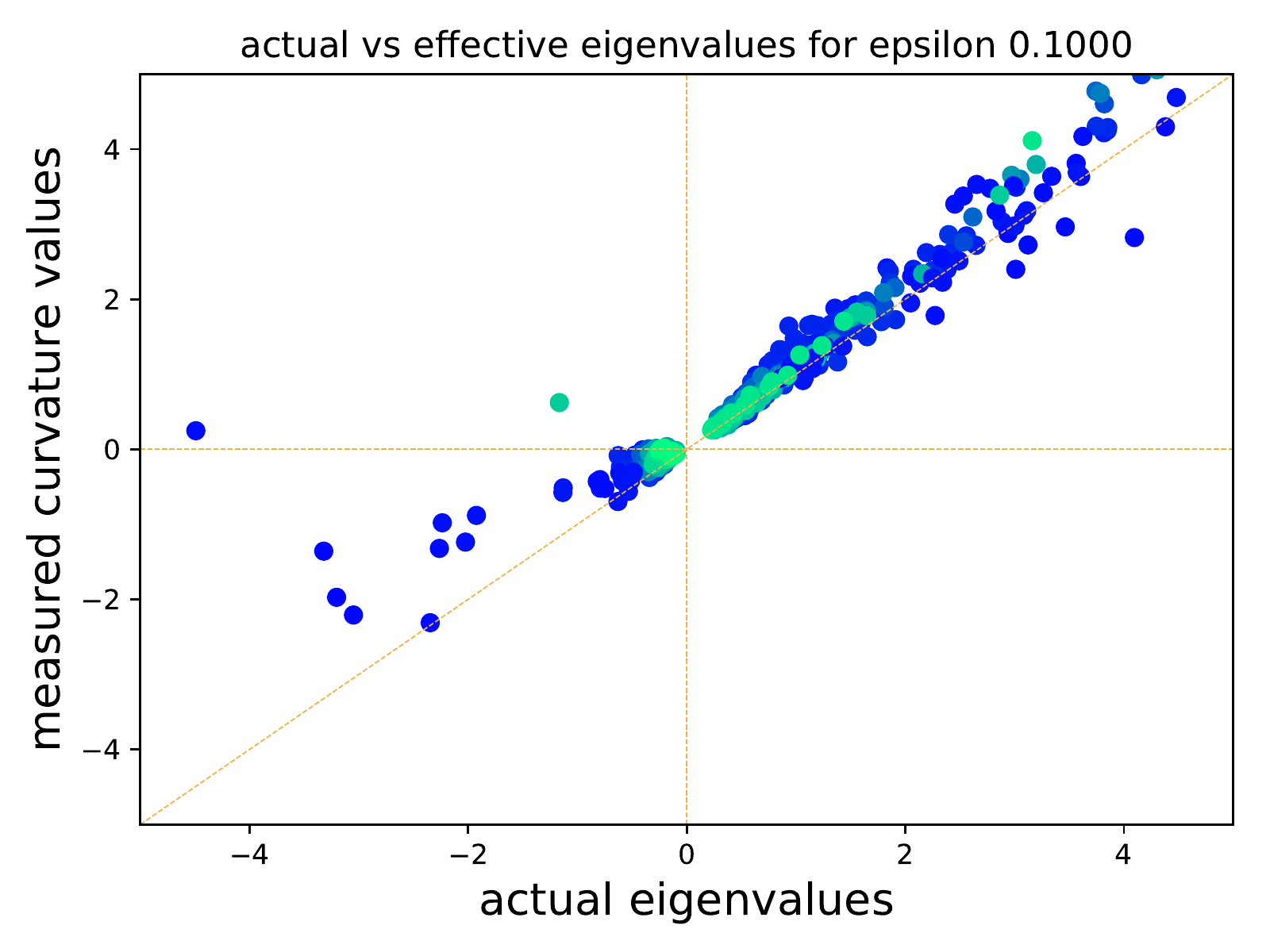}
\includegraphics[width=0.48\textwidth]{\paperroot/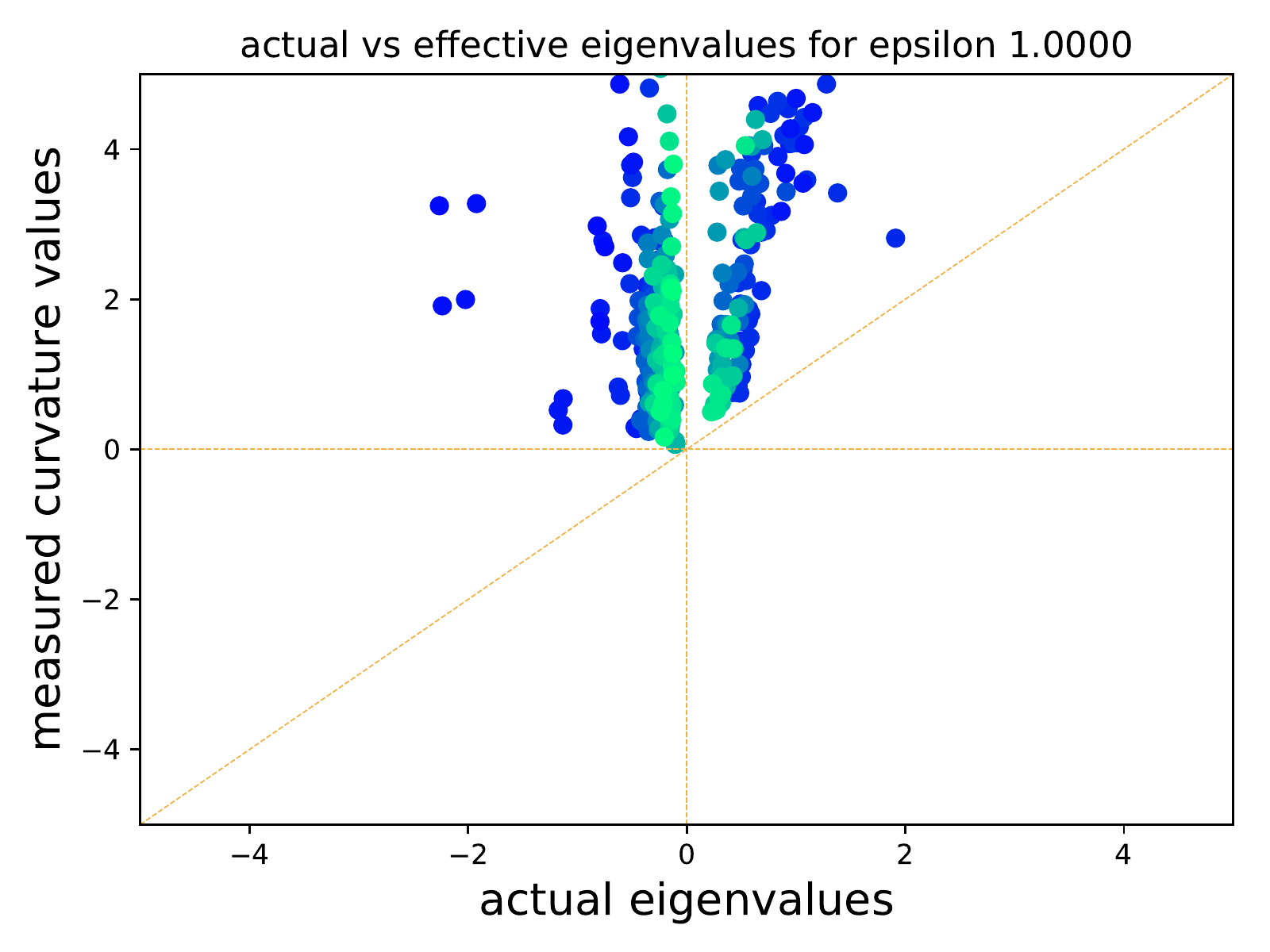}
\caption[Local vs global eigenvalues]{Comparison between the eigenvalues of the Hessian and the curvature computed on a larger scale ($|\alpha| = 0.1$ on the left and $|\alpha| = 1$ on the right). Blue points represent early stages of the optimization and green represent later stages. If the function were a true quadratic, all points would be on the $y=x$ line. We see that the global curvature is always larger than the local one but the effect is much more pronounced for the directions associated with negative eigenvalues.}
\label{neh-fig:local_vs_global_curvature}
\end{figure*}
The first question one may ask is whether the Hessian is a good representation of the local loss function. Indeed, blindly assuming that the function is quadratic when there is a direction of negative curvature would mean that the minimum is at infinity. As the unregularized loss is bounded below, any amount of $\ell_2$ regularization ensures that the loss $\mathcal{L}(\theta)$ goes to infinity as $\|\theta\|$ increases, proving the quadratic approximation is only valid locally. Cubic regularization is one way to address this issue by locally approximating the loss with a third-degree polynomial.

One way of understanding the spectral decomposition of the Hessian is by saying that the loss taken along each eigenvector $v_i$ is locally quadratic with curvature $\lambda_i$ where $\lambda_i$ is the eigenvalue associated with $v_i$. We thus compute the full eigenspectrum, i.e. all $(\lambda_i, v_i)$ pairs, of the Hessian during optimization, then compute the true loss along the direction $v_i$, i.e.
\begin{equation}
  \label{eqn:line_search}
  \mathcal{L}(\theta_t - \alpha \left[g(\theta)^\top v_i\right] v_i).
\end{equation}
We use the scaling factor $g(\theta)^\top v_i$ to represent what would happen should we move in the direction $v_i$ by computing the gradient then projecting it on that direction. Since we observed a common behaviour along the entire trajectory, we show here the results for an arbitrary iteration $(t=50)$.

From the previous section, we can expect the function to be approximately quadratic in the directions of large positive curvature but to contain higher-order components in the directions of negative curvature. Figure~\ref{neh-fig:measured_diff_vs_expolated_step} shows both the true loss (solid range curve) and the quadratic approximation (dashed blue/green curve) for $(\lambda_1, v_1)$, the eigenpair associated with the largest eigenvalue (top), and $(\lambda_d, v_d)$, the eigenpair associated with the smallest eigenvalue (bottom). We used $\alpha$ between -0.1 and +0.1 on the left and $\alpha$ between -1 and +1 on the right. The solid orange line is the true loss while the dotted green-blue line is the quadratic approximation. We observe a very different behaviour for the two eigenpairs. For the pair $(\lambda_1, v_1)$, while the quadratic approximation overestimates the change in loss for large values of $\alpha$, its quality remains acceptable. For the pair $(\lambda_d, v_d)$, however, the quadratic approximation is reasonable for small values of $\alpha$ but quickly falls apart as soon as $0.05 < |\alpha|$.

% We could also say that with the positive direction at least we are pointing in the correct curvature.]

To better quantify the difference between the local curvature as defined by the spectrum of the Hessian and a more global curvature as defined by a quadratic fit of the true loss along each direction, we first compute the second-order term $y_i$ in such a quadratic fit of the true loss for $\alpha \in [-0.1, 0.1]$ and for $\alpha \in [-1, 1]$. We then perform a scatterplot of each pair $(\lambda_i, y_i)$ with one point per direction defined by the $i$-th eigenvector where $\lambda_i$ is the corresponding eigenvalue and $y_i$ is the second-order term of the quadratic fit. The results can be seen in Figure~\ref{neh-fig:local_vs_global_curvature}. We see that all points are above the line, meaning the Hessian consistently underestimates the true curvature, and that the effect is more pronounced for directions of negative curvature. 

%To assess the impact of the cubic regularization, we fitted a third-order polynomial to the true losses of Figure~\ref{neh-fig:measured_diff_vs_expolated_step}.

In our experiments, the effective stepsize was much smaller than 0.1, hence corresponding to a scale where the quadratic approximation was correct. One might wonder if we could have used a larger stepsize with a proper treatment of these directions.

Since our tools for convex optimization fall apart in the directions of negative curvature, one might wonder what happens in these directions during the optimization. This is what we explore in the next section.

\subsection{Minimizing loss in directions of negative curvature}
\label{neh-sec:min_neg_curvature}

\subsubsection{Theoretically optimal step sizes}
\label{neh-appsec:optimal_step_sizes}

A strongly-convex loss function $f$ has a positive-definite Hessian matrix $H$ everywhere in the space, that is all its eigenvalues will be strictly greater than zero.

To perform an update with Newton's method, we update the parameters $\theta_t$ according to
\begin{equation*}
     \theta_{t+1} = \theta_{t} - \alpha H(\theta_t)^{-1} g(\theta_t)
\end{equation*}
where $g(\theta_t)$ is the gradient of $f(\theta)$ and $\alpha$ is the learning rate.

In the special case when $f(\theta)$ is quadratic, the Hessian is constant and we can use one Newton update with $\alpha=1$ to jump directly to the optimum. We can compute what that means in terms of the optimal step size to update $\theta$ along the direction of one of the eigenvector $v_i$.

Let $\left\{ (\lambda_1, v_1), \ldots, (\lambda_d, v_d)\right\}$ be the eigendecomposition of the Hessian matrix. If we project the gradient in the basis of eigenvectors, we get
\begin{equation*}
     g(\theta) = \sum_{i=1}^N \left[ g(\theta)^\top v_i\right] v_i.
\end{equation*}
Note that $H^{-1} v_i = \frac{1}{\lambda_i} v_i$, so we have that
\begin{equation*}
     H^{-1} g(\theta) = \sum_{i=1}^N \left[ g(\theta)^\top v_i\right] \frac{1}{\lambda_i} v_i.
\end{equation*}

Thus, when minimizing a strongly-convex quadratic function $f(\theta)$, the optimal step size along the direction of an eigenvector is given by
\begin{equation}
     \alpha^* = \argmin_\alpha \mathcal{L}\left( \theta - \alpha \left[ g(\theta)^\top v_i\right] v_i\right) = \frac{1}{\lambda_i}.
\end{equation}

If we are dealing with a strongly-convex function that is not quadratic, then the Hessian is not constant and we will need more than one Newton update to converge to the global minimum. However, we still obtain superlinear convergence in a ball around the optimum as the Hessian stabilizes.

In contrast, there is no result on the optimal step size for general functions. We can however measure optimal step sizes experimentally, as we do in the next section.

\subsubsection{Empirically optimal step sizes}

\begin{figure*}[ht!]
\includegraphics[width=.48\textwidth]{\paperroot/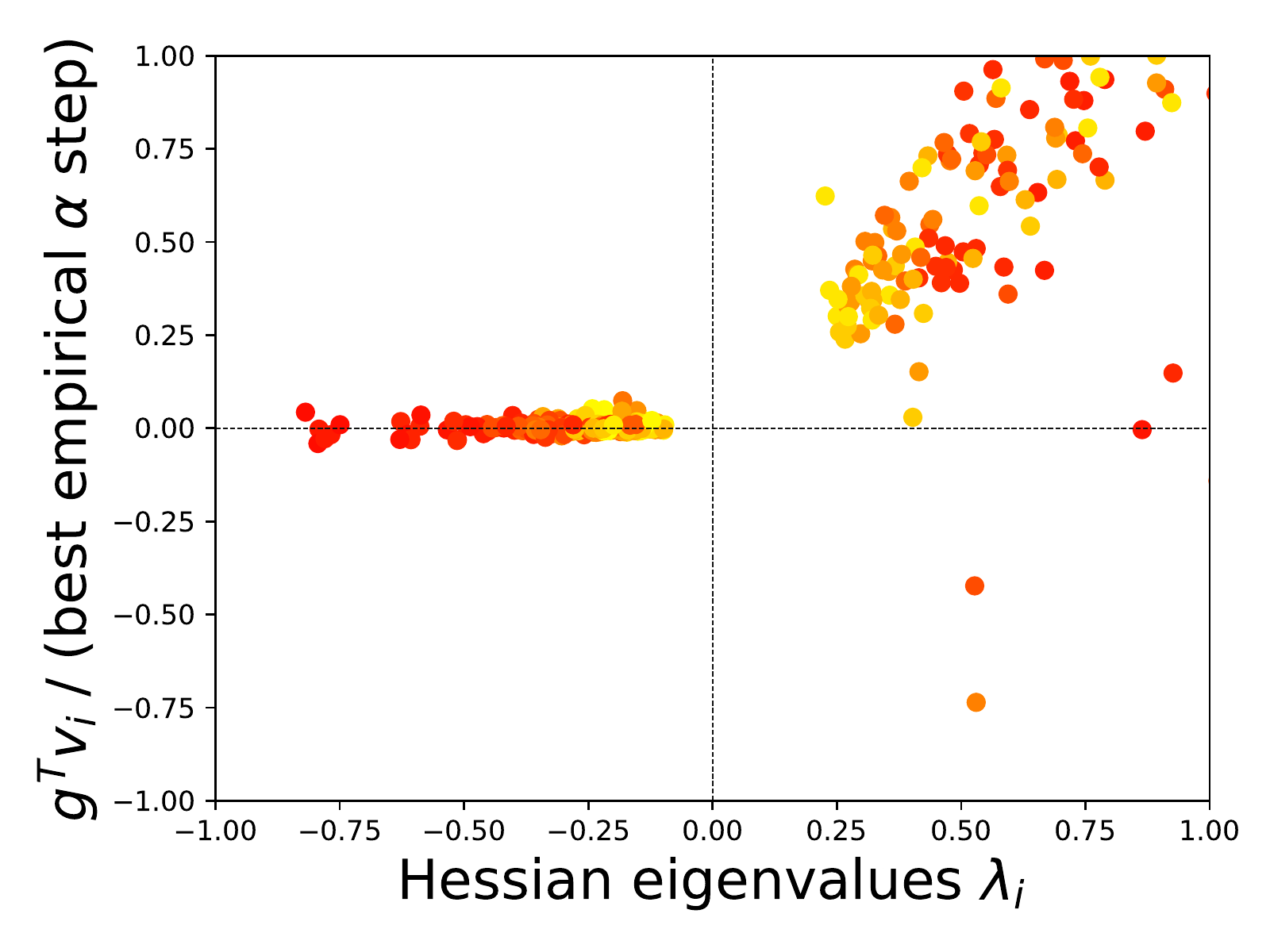}
\includegraphics[width=.48\textwidth]{\paperroot/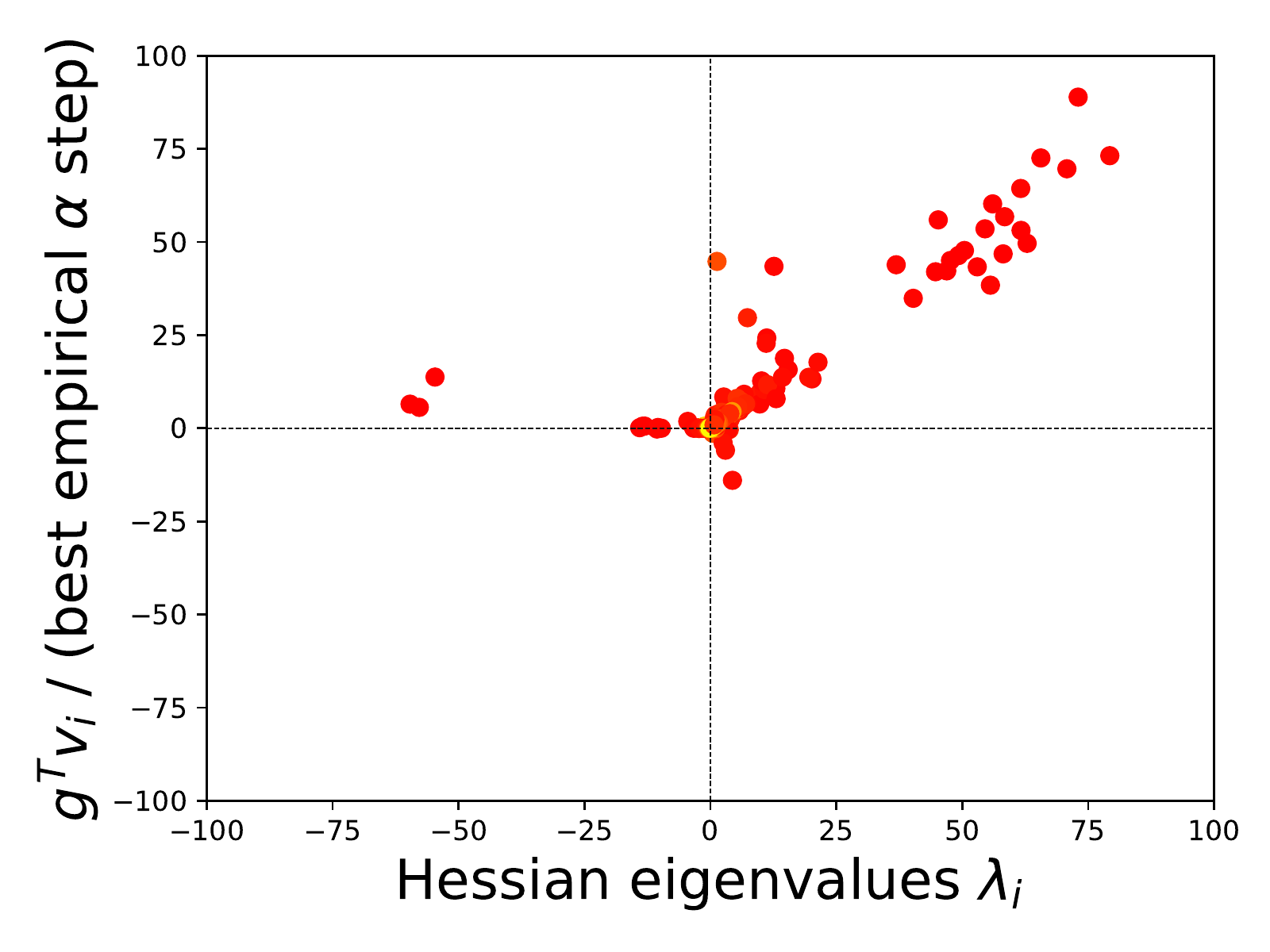}
\caption[Optimizing step sizes on different intervals]{Inverse of the optimal stepsize found through linesearch along each eigenvector $v_i$ as a function of the associated eigenvalue $\lambda_i$. We only show eigenvalues in $[-1, 1]$ (left) and in $[-100, 100]$ (right). Red points correspond to early stages in training and yellow points to later stages. For a true quadratic, all points would be on the $y=x$ line.}
\label{neh-fig:lambda_comparison}
\end{figure*}

While the quadratic approximation prescribes the use of a stepsize in the case of positive curvature, namely a stepsize of $1 / \rho$ when moving in a direction with curvature $\rho$, there is no prescribed value for the stepsize in the case of negative curvature. The main reason for this is that we know the quadratic approximation can only be trusted locally and so the update needs to be regularized toward a small value, as achieved by the addition of a third-order term in cubic regularization. some authors proposed heuristics. For instance, ~\citet{dauphin2014identifying} advocate for the use of a stepsize of $1 / |\rho|$.

We perform here an empirical study of the optimal stepsize for various curvatures. More specifically, we compute all eigenpairs $\{(\lambda_i, v_i)\}$ then perform a greedy line search in the direction of each $v_i$ to extract the optimal empirical stepsize $\alpha^*_i$, i.e.
\begin{align*}
    \alpha^*_i &= \arg\min_\alpha \mathcal{L}(\theta_t - \alpha (g^\top v_i)v_i) \; ,
\end{align*}
with $g$ the gradient at $\theta_t$.

Figure~\ref{neh-fig:lambda_comparison} shows that we indeed have $\alpha_i^* \approx 1 / \lambda_i$ for positive eigenvalues $\lambda_i$ but the relationship falls apart for negative eigenvalues. We also do not have $\alpha_i^* \approx 1 / |\lambda_i|$ but rather the optimal stepsize seems to be decorrelated from the eigenvalue. This result hints at the fact that we might need extra information to deal with negative curvature, for instance the third derivative.

\begin{figure*}[ht!]
\includegraphics[width=.48\textwidth]{\paperroot/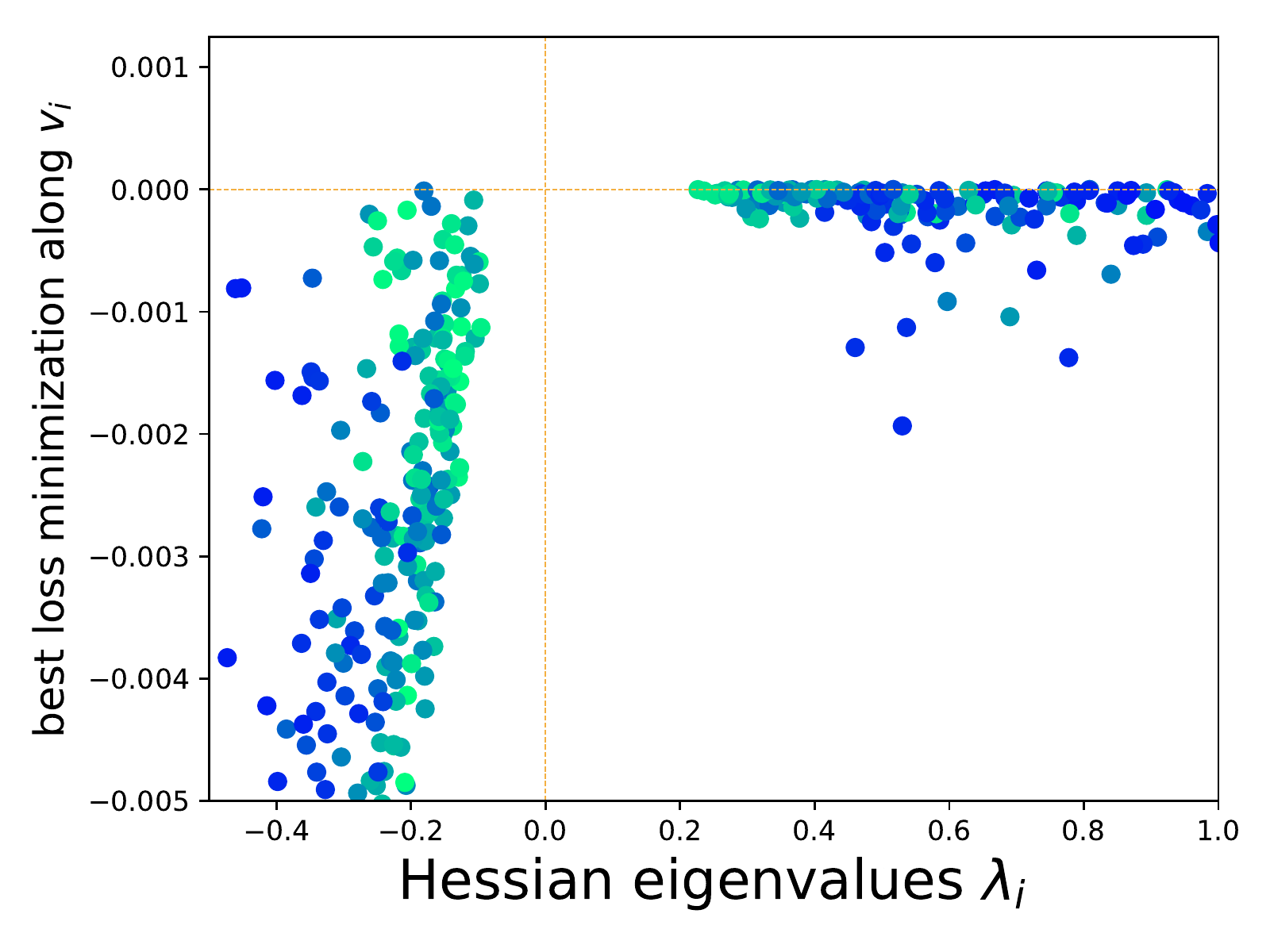}
\includegraphics[width=.48\textwidth]{\paperroot/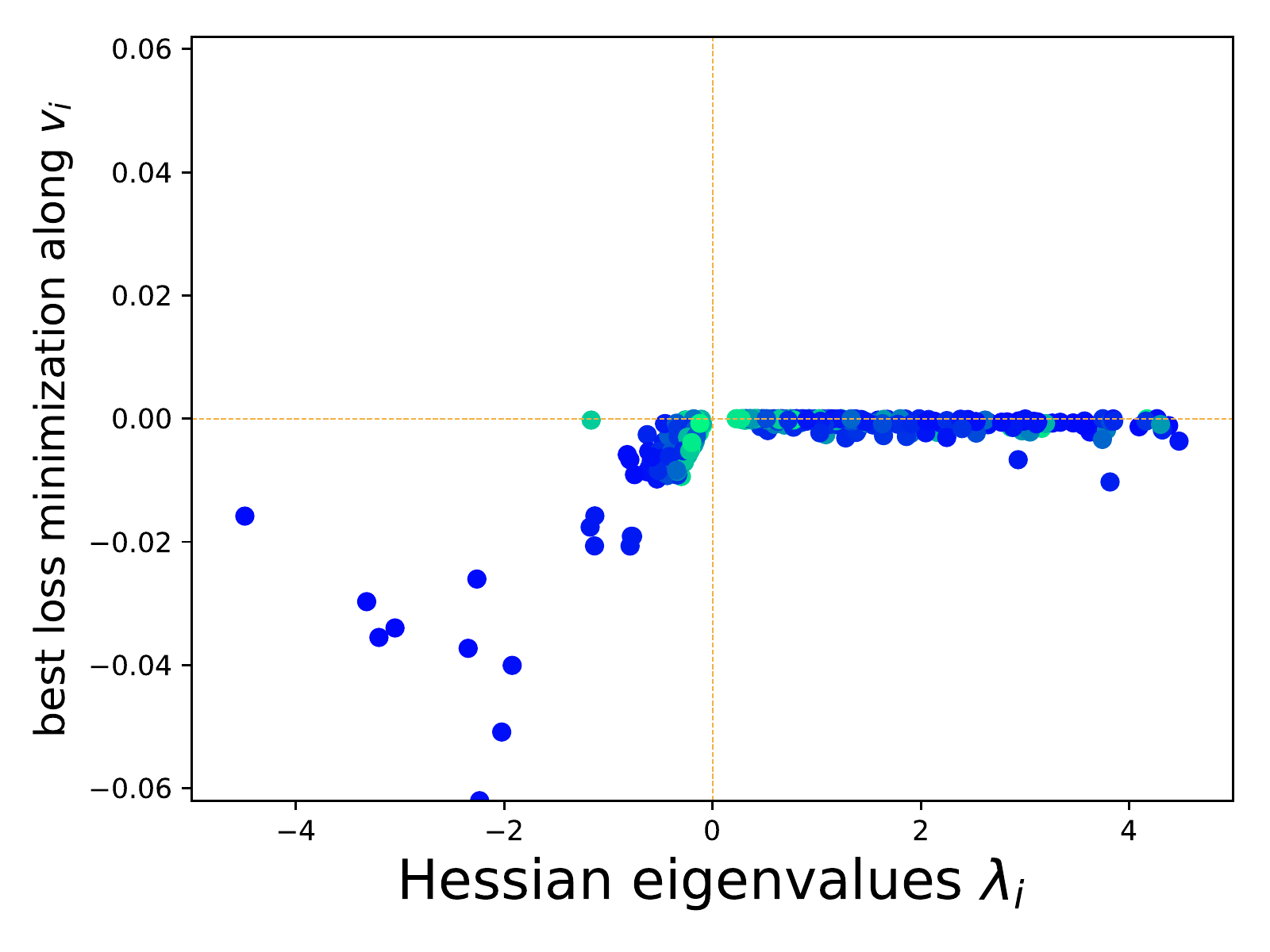}
\caption[Best loss improvements on different intervals]{Best loss improvement along each eigenvector $v_i$ as a function of the corresponding eigenvalue $\lambda_i$ for eigenvalues in $[-0.5, 0.5]$ (left) and in $[-5, 5]$ (right). Blue points correspond to early stages of training and green points to later stages.}
\label{neh-fig:actual_vs_loss_best}
\end{figure*}

A potential consequence of this poor estimate of the optimal stepsize is that we do not extract all the value from directions of negative curvature. To evaluate how much of the value lies in each direction, we compute the maximum loss improvement, i.e. the largest possible decrease, along each direction $v_i$ using the $\alpha_i^*$ computed before.

Figure \ref{neh-fig:actual_vs_loss_best} shows that the most improvement is obtained when optimizing in the directions of negative curvature. However, since there are far more directions of positive curvature than directions of negative curvature, the total gain is still larger for directions of positive curvature. We can also note that, in directions of positive curvature, blue points are on average lower than green points, meaning that there is less improvement to be had later in training, which is to be expected. In directions of negative curvature, however, the potential improvement remains stable throughout optimization, confirming that current optimizers do a poor job at exploiting directions of negative curvature. Since we are using numerical methods that report eigenvalues with the largest magnitude $|\lambda|$, those figures are missing more than 99.99\% of the eigenvalues with very small magnitude. This is why they do not have any points shown around the origin.

\section{Jacobian Vector Product}
\label{neh-appsec:jvp}

We now cover technical details regarding the computation of the spectrum of the Hessian.

With $d=3.3\times10^6$, the storage required to store the symmetric Hessian matrix with float32 coefficients is approximately $20$ terabytes,
which makes it close to impossible to store in RAM.
The task of computing all the $d$ eigenvalues is absolutely out of reach, but by using the ``Jacobian Vector Product'' trick ~\citep{townsend2017jvp}, along with Scipy ~\citep{jones2014scipy,lehoucq1998arpack}, we can compute the $k$ largest or smallest eigenpairs $(\lambda_i, v_i)$.

The Scipy library function \texttt{sparse.linalg.eigsh} is able to accept either a symmetric matrix, or a function that computes the product $v\mapsto H(\theta) v$. We define a Python function that makes many internal calls to Tensorflow to iterate over the whole training set (or a fixed subset thereof). We aggregate the results and return them. This enables a Scipy library function to make calls to Tensorflow without being aware of it.

Following again the notation \sectionname\ref{neh-sec:methodology}, we order the eigenvalues as $\lambda_1 \geq \lambda_2 \geq \ldots \geq \lambda_d$. They are all real-valued because the Hessian matrix is symmetric and contains only real coefficients.

We are mainly interested in the eigenvalues closest to $\pm \infty$, so we define the following notation to refer to the $k$ most extreme eigenpairs on each side.
\begin{eqnarray*}
     \textrm{LA}(k) & = & \left\{ (\lambda_1, v_1), \ldots, (\lambda_k, v_k)\right\} \\
     \textrm{SA}(k) & = & \left\{ (\lambda_{d-k+1}, v_{d-k+1}), \ldots, (\lambda_d, v_d)\right\}.
\end{eqnarray*}

Note that the costs of computing those sets depends a lot of the magnitude of the eigenvalues. In practice we observed that the LA eigenvalues have a much larger magnitude than the SA. %(see Figure \ref{neh-sec:eig_imperfect_eigvals_100}).
This leads to the task of computing $\textrm{LA}(20)$ being much cheaper than $\textrm{SA}(3)$, despite the fact that it involves more eigenvalues.

For reasons of computational costs, we resorted to using a fixed subset of the training set when we performed the eigendecompositions (done after training).

\section{Conclusion and future work}
Building on previous analyses of the Hessian of deep networks, we studied the quality of the quadratic approximation as well as the impact of directions of negative curvatures. We emphasized the importance of handling them differently than the directions of positive curvature.

In particular, we assessed how quickly the quadratic approximation falls apart and how the ``global'' curvature differs from the ``local'' one. We also provide an empirical answer to the question of the optimal stepsize in directions of negative curvature, further showing how a better treatment of these directions could lead to further gains in the training loss.

We hope our study provides insights into the specific nonconvexity of the loss of deep networks and will ultimately guide the design of tailored optimizers.

\subsubsection*{Acknowledgments}
We thank Bart van Merri\"enboer for fruitful discussions about optimization and the problem of saddle points. We thank Ying Xiao for initial discussions about his preliminary work on studying eigenvalues, and for providing his code to quickly get the ``Jacobian vector product'' trick working.

% \bibliography{negative_eigenvalues}
\bibliography{arXiv}

\begin{thebibliography}{19}
\providecommand{\natexlab}[1]{#1}
\providecommand{\url}[1]{\texttt{#1}}
\expandafter\ifx\csname urlstyle\endcsname\relax
  \providecommand{\doi}[1]{doi: #1}\else
  \providecommand{\doi}{doi: \begingroup \urlstyle{rm}\Url}\fi

\bibitem[Adams et~al.(2018)Adams, Pennington, Johnson, Smith, Ovadia, Patton,
  and Saunderson]{adams2018estimating}
Ryan~P Adams, Jeffrey Pennington, Matthew~J Johnson, Jamie Smith, Yaniv Ovadia,
  Brian Patton, and James Saunderson.
\newblock Estimating the spectral density of large implicit matrices.
\newblock \emph{arXiv preprint arXiv:1802.03451}, 2018.

\bibitem[Allen-Zhu(2017)]{allen2017natasha}
Zeyuan Allen-Zhu.
\newblock Natasha 2: Faster non-convex optimization than sgd.
\newblock \emph{arXiv preprint arXiv:1708.08694}, 2017.

\bibitem[Bottou et~al.(2018)Bottou, Curtis, and
  Nocedal]{bottou2018optimization}
L{\'e}on Bottou, Frank~E Curtis, and Jorge Nocedal.
\newblock Optimization methods for large-scale machine learning.
\newblock \emph{SIAM Review}, 60\penalty0 (2):\penalty0 223--311, 2018.

\bibitem[Choromanska et~al.(2015)Choromanska, Henaff, Mathieu, Arous, and
  LeCun]{choromanska2015loss}
Anna Choromanska, Mikael Henaff, Michael Mathieu, G{\'e}rard~Ben Arous, and
  Yann LeCun.
\newblock The loss surfaces of multilayer networks.
\newblock In \emph{Artificial Intelligence and Statistics}, pp.\  192--204,
  2015.

\bibitem[Dauphin et~al.(2014)Dauphin, Pascanu, Gulcehre, Cho, Ganguli, and
  Bengio]{dauphin2014identifying}
Yann~N Dauphin, Razvan Pascanu, Caglar Gulcehre, Kyunghyun Cho, Surya Ganguli,
  and Yoshua Bengio.
\newblock Identifying and attacking the saddle point problem in
  high-dimensional non-convex optimization.
\newblock In \emph{Advances in neural information processing systems}, pp.\
  2933--2941, 2014.

\bibitem[De~Sa et~al.(2014)De~Sa, Olukotun, and R{\'e}]{de2014global}
Christopher De~Sa, Kunle Olukotun, and Christopher R{\'e}.
\newblock Global convergence of stochastic gradient descent for some non-convex
  matrix problems.
\newblock \emph{arXiv preprint arXiv:1411.1134}, 2014.

\bibitem[Guy Gur-Ari \& Dyer(2018)Guy Gur-Ari and Dyer]{gurari2018tiny}
Daniel A.~Roberts Guy Gur-Ari and Ethan Dyer.
\newblock Gradient descent happens in a tiny subspace.
\newblock \emph{arXiv preprint arXiv:1812.04754}, 2018.

\bibitem[He et~al.(2016)He, Zhang, Ren, and Sun]{he2016deep}
Kaiming He, Xiangyu Zhang, Shaoqing Ren, and Jian Sun.
\newblock Deep residual learning for image recognition.
\newblock In \emph{Proceedings of the IEEE conference on computer vision and
  pattern recognition}, pp.\  770--778, 2016.

\bibitem[Hinton et~al.(2012)Hinton, Srivastava, and Swersky]{hinton2012neural}
Geoffrey Hinton, Nitish Srivastava, and Kevin Swersky.
\newblock Neural networks for machine learning lecture 6a overview of
  mini-batch gradient descent, 2012.

\bibitem[Jones et~al.(2014)Jones, Oliphant, and Peterson]{jones2014scipy}
Eric Jones, Travis Oliphant, and Pearu Peterson.
\newblock $\{$SciPy$\}$: open source scientific tools for $\{$Python$\}$, 2014.

\bibitem[LeCun \& Cortes(1998)LeCun and Cortes]{lecun1998mnist}
Yann LeCun and Corinna Cortes.
\newblock The mnist database of handwritten digits, 1998.

\bibitem[LeCun et~al.(1989)LeCun, Boser, Denker, Henderson, Howard, Hubbard,
  and Jackel]{lecun1989backpropagation}
Yann LeCun, Bernhard Boser, John~S Denker, Donnie Henderson, Richard~E Howard,
  Wayne Hubbard, and Lawrence~D Jackel.
\newblock Backpropagation applied to handwritten zip code recognition.
\newblock \emph{Neural computation}, 1\penalty0 (4):\penalty0 541--551, 1989.

\bibitem[Lehoucq et~al.(1998)Lehoucq, Sorensen, and Yang]{lehoucq1998arpack}
Richard~B Lehoucq, Danny~C Sorensen, and Chao Yang.
\newblock \emph{ARPACK users' guide: solution of large-scale eigenvalue
  problems with implicitly restarted Arnoldi methods}.
\newblock SIAM, 1998.

\bibitem[Nesterov \& Polyak(2006)Nesterov and Polyak]{nesterov2006cubic}
Yurii Nesterov and Boris~T Polyak.
\newblock Cubic regularization of newton method and its global performance.
\newblock \emph{Mathematical Programming}, 108\penalty0 (1):\penalty0 177--205,
  2006.

\bibitem[Papyan(2018)]{papyan2018eigenvalues}
Vardan Papyan.
\newblock The full spectrum of deep net hessians at scale: Dynamics with sample
  size.
\newblock \emph{arXiv preprint arXiv:1811.07062}, 2018.

\bibitem[Sagun et~al.(2016)Sagun, Bottou, and LeCun]{sagun2016eigenvalues}
Levent Sagun, Leon Bottou, and Yann LeCun.
\newblock Eigenvalues of the hessian in deep learning: Singularity and beyond.
\newblock \emph{arXiv preprint arXiv:1611.07476}, 2016.

\bibitem[Townsend(2017)]{townsend2017jvp}
Jamie Townsend.
\newblock {A new trick for calculating Jacobian vector products}.
\newblock \url{https://j-towns.github.io/2017/06/12/A-new-trick.html}, 2017.
\newblock [Online; accessed 20-Jan-2018].

\bibitem[Tripuraneni et~al.(2017)Tripuraneni, Stern, Jin, Regier, and
  Jordan]{tripuraneni2017stochastic}
Nilesh Tripuraneni, Mitchell Stern, Chi Jin, Jeffrey Regier, and Michael~I
  Jordan.
\newblock Stochastic cubic regularization for fast nonconvex optimization.
\newblock \emph{arXiv preprint arXiv:1711.02838}, 2017.

\bibitem[Zhang et~al.(2016)Zhang, Bengio, Hardt, Recht, and
  Vinyals]{zhang2016understanding}
Chiyuan Zhang, Samy Bengio, Moritz Hardt, Benjamin Recht, and Oriol Vinyals.
\newblock Understanding deep learning requires rethinking generalization.
\newblock \emph{arXiv preprint arXiv:1611.03530}, 2016.

\end{thebibliography}
\bibliographystyle{iclr2018_workshop}

\appendix

\section{Optimal step sizes}
\label{neh-appsec:optimal_step_sizes}

A strictly-convex loss function $f(\theta)$ has a positive-definite Hessian matrix $H(\theta)$ for all values of $\theta$. That is, all its eigenvalues will be strictly greater than zero.

To perform an update with Newton's method, we update the parameters $\theta_t$ according to
\begin{equation*}
     \theta_{t+1} = \theta_{t} - \alpha H(\theta_t)^{-1} g(\theta_t)
\end{equation*}
where $g(\theta_t)$ is the gradient of $f(\theta)$ and $\alpha$ is the learning rate.

In the special case when $f(\theta)$ is quadratic, the Hessian is constant and we can use one Newton update with $\alpha=1$ to jump directly to the optimum. We can compute what that means in terms of the optimal step size to update $\theta$ along the direction of one of the eigenvector $v_i$.

Let $\left\{ (\lambda_1, v_1), \ldots, (\lambda_d, v_d)\right\}$ be the eigendecomposition of the Hessian matrix. If we project the gradient in the basis of eigenvectors, we get
\begin{equation*}
     g(\theta) = \sum_{i=1}^N \left[ g(\theta)^\top v_i\right] v_i.
\end{equation*}
Note that $H^{-1} v_i = \frac{1}{\lambda_i} v_i$, so we have that
\begin{equation*}
     H^{-1} g(\theta) = \sum_{i=1}^N \left[ g(\theta)^\top v_i\right] \frac{1}{\lambda_i} v_i.
\end{equation*}

Thus, when minimizing a strictly-convex quadratic function $f(\theta)$, the optimal step size along the direction of an eigenvector is given by
\begin{equation}
     \alpha^* = \argmin_\alpha \mathcal{L}\left( \theta - \alpha \left[ g(\theta)^\top v_i\right] v_i\right) = \frac{1}{\lambda_i}.
\end{equation}

If we are dealing with a strictly-convex function that is not quadratic, then the Hessian is not constant and we will need more than one Newton update to converge to the global minimum. We can still hope that a step size of $1/\lambda_i$ would be a good value to use.

With a deep neural network, we no longer have any guarantees. We can still measure optimal step sizes experimentally, which is what we have done in \sectionname~\ref{neh-sec:min_neg_curvature}. We saw in \figurename~\ref{neh-fig:lambda_comparison} that the optimal step sizes in directions $v_i$ of positive curvature matched rather well with the value of $1/\lambda_i$.
It has been suggested in ~\citet{dauphin2014identifying} that in directions of negative curvature, the optimal step size could be $1/\left|\lambda_i\right|$, but our empirical results are much larger than that. Again, we have to keep in mind that a general theory cannot be extrapolated from only one model and one dataset.

% % \ref{neh-fig:actual_vs_loss_best}

% %On the left side of the plot, we could have expected to find $1/\alpha^* = |\lambda|$ based on ~\citet{dauphin2014identifying}. Instead we find that the optimal step sizes are empirically much larger than that. This means that we are probably too conservative in practice and we should be more aggressive about step sizes in directions of negative curvature.

\section{On estimating the Hessian}

Given that the full Hessian matrix has more than $10^{13}$ coefficients, and that the entire training set has $50000*28^2$ coefficients, we might be concerned about whether the value of the Hessian is possible to estimate statistically.

In a way, much like the loss $\mathcal{L}(\theta) = \sum_{n=1}^N \mathcal{L}_\theta(x_i, y_i)$ is an exact quantity defined over the whole training set, the Hessian is the same. The notion of an estimator variance would come into play if we estimated $H(\theta)$ from a minibatch instead.

Given the computational costs of evaluating $\mathcal{L}(\theta)$ and $H(\theta)$ on the whole training set every time that the Scipy function \texttt{scipy.sparse.linalg.eigsh} wants us to evaluate the Jacobian vector product, we tried to see if it was possible to get away with only using 5\% of the training set for that purpose. That 5\% has to always contain the same samples, or otherwise we violate assumptions made by Scipy (in a way similar to how the usual quicksort implementation would fail if comparisons were no longer deterministic).

Now $H_{5\%}(\theta)$ is an estimator of $H(\theta)$, and we have verified experimentally that the first elements of the eigenspectrum of those two matrices are close enough for the purposes of our analysis. We did this by comparing $\textrm{LA}(10)$ and $\textrm{SA}(10)$ in both cases, checking the differences between eigenvalues and the angles between the eigenvectors.
It was important to check to see if we would have numerical instabilities with a regime using less data.

\section{Suggestion for new optimization method}
\label{neh-appsec:newmethod}

Considerable work was required for us to extract negative eigenvalues for every checkpoint of training. This is not a practical thing to do during training, so we want to introduce here the idea of keeping a running approximation of the smallest eigenvector of the Hessian.

We know that the Jacobian vector product $H(\theta)v$ can be evaluated on a minibatch at the same time that we compute the gradient. Some people report an overhead of $~4\times$ the computational costs, but we have not measured any benchmarks in that regards.

The smallest eigenvector is a unit vector $v$ that minimizes the value of $m(v) = v^\top H(\theta)v$. This is a quadratic in the coefficients of $v$ (along with a constraint on the norm of $v$), and it's something that we can minimize using a method similar to SGD. We can easily see that $\nabla_v m(v) = 2 H(\theta)v$, so we can minimize simultaneously $m(v)$ and the usual model loss $\mathcal{L}(\theta)$. This means that we can keep a running estimate $(\tilde{\lambda}, \tilde{v})$ of $(\lambda_d, v_d)$, and we can alternate between one update to $\theta$ with the usual RMSProp/Adam optimizer, and then one update in the direction of $\left[ g(\theta)^\top \tilde{v}\right] \tilde{v}$. Different learning rates could be used for those updates. If we wanted to minimize the overhead, we could also scale back to do those updates less frequently.

This is not something that we have tried in practice, but it would be the most direct way to implement a training method based on the ideas of this paper.

\end{document}